\documentclass[submission,copyright,creativecommons]{eptcs}
\providecommand{\event}{ICLP 2019 DC} % Name of the event you are submitting to
\usepackage{breakurl}             % Not needed if you use pdflatex only.
\usepackage{underscore}           % Only needed if you use pdflatex.

\title{Reasoning about Qualitative Direction and Distance between Extended Objects using Answer Set Programming}  %An Example of a Paper\\ with a Rather Large Title-to-Content Ratio}
\author{Yusuf Izmirlioglu \thanks{This thesis is supervised by Esra Erdem.}
\institute{Computer Science and Engineering\\ Sabanci University, Turkey}   % \thanks{A fine university.}
\email{yizmirlioglu@sabanciuniv.edu}
}
\def\titlerunning{Reasoning about Qualitative Direction and Distance using ASP}
\def\authorrunning{Y. Izmirlioglu}

\usepackage{times}
\usepackage{helvet}
\usepackage{courier}
\usepackage{url}
\usepackage{graphicx}
\usepackage{amsmath,amssymb,amsthm}
\usepackage{latexsym}
\usepackage{tabulary}
\usepackage{float}

\usepackage{booktabs} % for much better looking tables
\usepackage{array} % for better arrays (eg matrices) in maths
\usepackage{paralist} % very flexible & customisable lists (eg. enumerate/itemize, etc.)
\usepackage{verbatim} % adds environment for commenting out blocks of text & for better verbatim
\usepackage{subfig} % make it possible to include more than one captioned figure/table in a single float

\usepackage[titles,subfigure]{tocloft} % Alter the style of the Table of Contents
\usepackage{multicol}
\usepackage{listings}

\def\ba{\begin{array}}
	\def\ea{\end{array}}
\def\bi{\begin{itemize}}
	\def\ei{\end{itemize}}
\def\be{\begin{enumerate}}
	\def\ee{\end{enumerate}}
\def\beq{\begin{equation}}
\def\eeq#1{\label{#1}\end{equation}}

\def\ii#1{\hbox{\sl #1\/}}

\def\eqs{\,{=}\,}

\usepackage{todonotes}

\begin{document}
\maketitle

\begin{abstract}
In this thesis, we introduce a novel formal framework to represent and reason about qualitative direction and distance relations between extended objects using Answer Set Programming (ASP). We take Cardinal Directional Calculus (CDC) as a starting point and extend CDC with new sorts of constraints which involve defaults, preferences and negation. We call this extended version as nCDC. Then we further extend nCDC by augmenting qualitative distance relation and name this extension as nCDC+. For CDC, nCDC, nCDC+, we introduce an ASP-based general framework to solve consistency checking problems, address composition and inversion of qualitative spatial relations, infer unknown or missing relations between objects, and find a suitable configuration of objects which fulfills a given inquiry.
\end{abstract}

\section{Introduction}

%In my thesis, I study reasoning about qualitative directions and distance using Answer Set Programming. 
%%The topic of my thesis is Qualitative Spatial Reasoning about Cardinal Directions and Distance using Answer Set Programming. 
%I also perform an application to the robotic problem of geometric placement of multiple movable objects.  % in the scope of the thesis.

%The motivation for this research is that 
Spatial representation and reasoning is an essential component of geographical information systems, cognitive robotics, spatial databases, document interpretation and digital forensics.  % and robotic navigation.
Many tasks in these areas, such as satellite image retrieval, navigation of a robot to a destination, describing the location of an object, constructing digital maps involve dealing with spatial properties of the objects and the environment.

%Spatial reasoning and representation about locations of objects is an essential aspect of artificial intelligence and robotic applications in motion planning, navigation and manipulation.

In some domains (e.g., exploration of an unknown territory), qualitative models are more suitable for representing and reasoning about spatial relations because quantitative data may not always be available due to uncertainty or incomplete knowledge. 
In cognitive systems, spatial information obtained through perception might be coarse or imperfect.
Even if quantitative data is available, in some circumstances agents may prefer to use qualitative terms for the sake
of sociable and understandable communication.
%The reason is that 
For instance, humans express orientation and distance in words like \textit{left}, \textit{right}, \textit{front}, \textit{back}, \textit{north}, \textit{near}.
Naval, air and space navigation typicaly involve geographical directions such as \textit{south}, \textit{east}, \textit{northwest}.
Although qualitative terms have less resolution in geometry than their quantitative counterparts, it is easier
for people to communicate using them. %process and communicate them.
%In fact qualitative explanations can provide more information and help people to focus as it eliminates unnecessary details.
%For an average person, 
It is more eloquent to say ``The library is in front of the theater, near the cafeteria"
rather than ``The library is at 38.6 latitude and 27.1 longitude.  %385 meters and 30 degrees from here".
This explains why driving instructions in GPS system are conducted in daily language.

%Similarly "Singapore is at 2.34 latitude, 165.7 longitude and 48.2 km2" does not make much sense whereas "Singapore is at the southeast of Malaysia and very close to equator; its land area is almost the same as area of Philadelphia." is cognitively much more meaningful.

\begin{figure}[h]
\centering
\includegraphics[width = 0.5\linewidth]{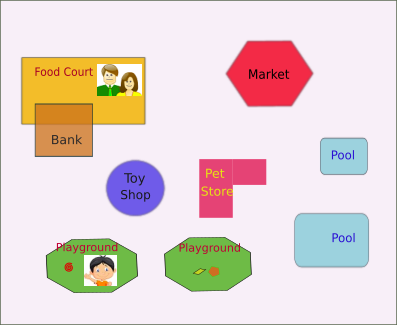}
\caption{Missing child scenario}
\label{fig:scenario}
\end{figure}

As an illustrative scenario (depicted in Figure~\ref{fig:scenario}), suppose that a robot is assisting a parent to find her missing child in a shopping mall that is not completely known to the robot nor to the parents. 
The robot has received some sightings of the child (e.g., ``to the south of Store A").
This information will be useful if the robot can understand the relative location of the child described qualitatively, figure out where the child might be, based on such qualitative direction constraints, and describe qualitatively in which direction (e.g., ``to the north") the parents should search for their child.
%In another scenario, a robot may not have quantitative description of an environment (e.g., after a disaster, as in search and rescue), and should be able to understand a human if she provides information by directions, reason about the qualitative constraints provided by the humans, and describe qualitatively the outcome of its reasoning.

\begin{figure}[h]
\centering
\includegraphics[width = 0.5\linewidth]{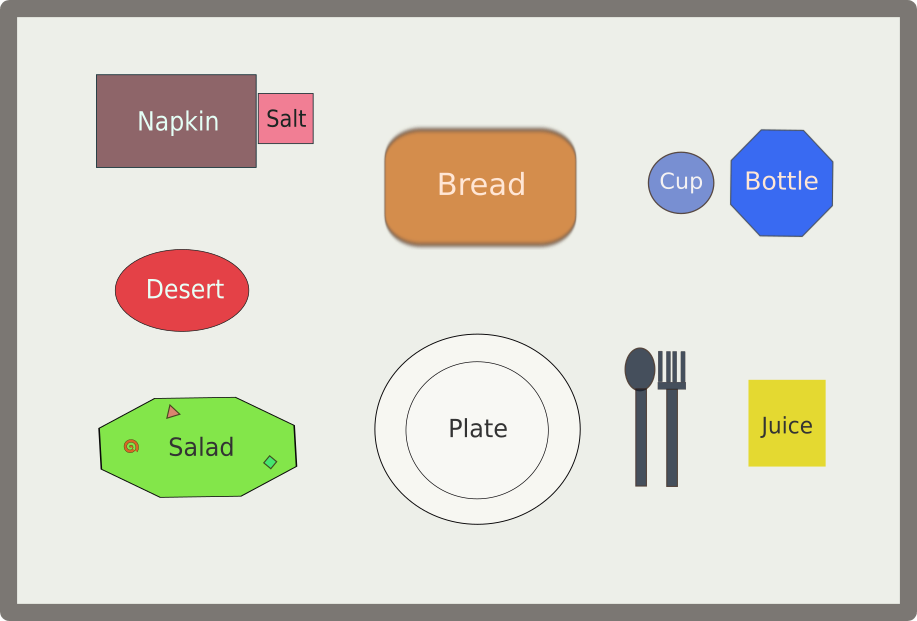}
\caption{Kitchen table scenario}
\label{fig:scenario2}
\end{figure}

%As an another example, 
In another scenario depicted in Figure~\ref{fig:scenario2}, a human user requests a service robot to prepare the kitchen table.
The commonsense knowledge for a well-set table can be described using statements such as ``The plate is in the middle of the table", ``Spoon is on the right and very near to the plate", ``Desert is between napkin and salad", ``Salad is on the left and near to the plate", ``Salt is adjacent to the napkin which is near to the top border", 
%When the user is unsure or flexible about position of some objects, he might say "The cup is to be located to the right or back and very near to the bottle."
``The cup is on the right or back and very near to the bottle."
%Or instead of a strict constraint, he can also reveal a preference like 
The human might express his preferences as well: ``It is better if the juice is placed on the right side of the plate, not far and not very near to it".
To set up the table, the robot should possess representation of these qualitative direction and distance relations between objects in order to understand the human and infer the setting.
It is also beneficial if the robot can utilize commonsense knowledge to enhance the arrangement, %. As an example he should know that 
for example ``by default the fork is placed next to the spoon".  % situated adjacent to the spoon". 

%Orientation is a fundamental aspect of space that we focus in the thesis.
%%In this thesis, we focus on orientation as it is a fundamental aspect of space.
%We are also aware the fact that combining multiple aspects of space such as direction, distance, size, shape has more potential for practical applications in geographical imaging, video processing, robotic navigation and manipulation since expressiveness is increased.
%Besides, people tend to describe position of a landmark or object not just by orientation but possibly with distance and size.
%% People express location of a landmark using relative direction with respect to his own position or another place.

With these motivations, our objective in this thesis is to develop a general framework to represent constraints, commonsense knowledge, preferences about qualitative directions and distances, to check consistency of these information, and to infer unknown spatial relations.
We propose to develop this framework using Answer Set Programming (ASP).

\section{Related Literature}

Beginning with the seminal work of Allen on Interval Algebra~\cite{Allen83}, 
a multitude of qualitative calculi have been proposed in the literature focusing on 
different aspects of space. 
Some of these formalisms focus on topology (DIR9~\cite{egenhofer1990categorizing}, RCC8~\cite{cohn1997qualitative}), 
direction (cone and projection based~\cite{Frank91}, 
LR~\cite{ligozat1993qualitative}, Double-cross~\cite{Freksa1992}, Dipole~\cite{moratz2000qualitative}, SV~\cite{LeeRW13},
OPRA~\cite{moratz2005relative}, 
Rectangle Algebra~\cite{balbiani1998model,BalbianiCC99}, 
Cardinal Directional Calculus~\cite{Goyal1997, SkiadKoub2004}), 
distance~\cite{zimmermann1996qualitative,monferrer1996enhancing,falomir2013qualitative,guesgen2002reasoning}, 
size~\cite{Frank91}, and shape~\cite{dugat1999qualitative,gottfried2005global,van2005double,museros2004qualitative,dorr2014qualitative}. 
An overview of qualitative spatial and temporal calculus can be found in recent surveys~\cite{cohn2008qualitative,chen2015survey,dylla2017survey}.

As for direction, point objects~\cite{Frank91, moratz2005relative, LeeRW13}, 
line segments and ternary relations~\cite{Freksa1992, moratz2002qualitative}, and 
extended regions on the plane~\cite{BalbianiCC99, Goyal1997} have been examined. 
These formalisms are designed for point objects; in this thesis we consider extended objects.

Rectangle Algebra~\cite{BalbianiCC99} and 
Cardinal Directional Calculus~\cite{Goyal1997, SkiadKoub2004, SkiadKoub2005} 
are widely used for reasoning about directions between extended objects on the plane. 
%are more powerful.
%RA is 2-dimensional externsion of Allen's Interval Algebra whose sides.. An RA relation is identified by a pair of 
Rectangle Algebra (RA) is an extension of Allen's Interval Algebra into 2-dimension. Objects are rectangles whose sides are parallel to axes of reference frame. An RA relation is identified by a pair of interval relation between sides of rectangles in horizontal and vertical axis. 
%~\cite{Goyal1997} developed Direction Relation Matrix where the space is divided into 9 tiles and
In Direction Relation Matrix ~\cite{Goyal1997}, the space is divided into 9 tiles and  
the direction of the target object is represented by its intersection 
with the tiles in a 3x3 matrix. 
A more formal model was adapted by Skiadopoulos and Koubarakis~\cite{SkiadKoub2004,SkiadKoub2005} in a manner that lower 
dimensional parts (points and lines) do not alter directional relations.
Its new form is named Cardinal Directional Calculus (CDC). 
In this thesis, our studies regarding directions is based on CDC.
Consistency checking in CDC and its computational complexity 
have been investigated in subsequent research~\cite{Liuthesis2013,LiLiu2011,Liuetal2010,Navarreteetal2007,SkiadKoub2004,SkiadKoub2005,Zhangetal2008}. 
%Subsequent research~\cite{Liuthesis2013,LiLiu2011,Liuetal2010,Navarreteetal2007,SkiadKoub2004,SkiadKoub2005,Zhangetal2008} have studied consistency checking in CDC and its computational complexity. 
Polynomial time complexity fragments of the problem have been identified~\cite{Liuthesis2013,Liuetal2010,Navarreteetal2007,Zhangetal2008} and algorithms have been presented for them. 
Although consistency checking problem is proven to be NP-complete in general~\cite{Liuthesis2013,LiLiu2011,Liuetal2010,SkiadKoub2005},
no solution method exists for these intractable problems in the literature.

There are also calculi that integrate different aspects such as topology and orientation~\cite{hotzcombining}, 
orientation and distance~\cite{clementini1997qualitative,moratz2002qualitative,moratz2003spatial,moratz2012spatial}, 
topology and size~\cite{gerevini2002combining},
topology, size and distance~\cite{brageul2007model}. 
These formalisms consider solely point objects for describing combined spatial relations.
We aim to construct a formal framework for reasoning about directions and distance for extended planar objects in the thesis.
We consider qualitative distance which includes symbolic relations with adjustable granularity.   % arbitrary 

In the literature, qualitative direction and distance are combined to define a \textit{qualitative position}. 
Symbolic binary distance relation is augmented into cone-based cardinal directions with granularity $k$ ~\cite{clementini1997qualitative}.  %  /arbitrary granularity
This model with four cone-based cardinal directions (\textit{north, south, east, west}) and four interval-based distance relations has been further investigated~\cite{hong1995robustness}.

As for other formalisms that combine direction with distance, in one study~\cite{zimmermann1996qualitative} LR calculus is enriched with a comparative distance relation. 
Orientation of a point $c$ is identified with respect to the directed line segment across the two reference objects $a,b$ and denoted by $ab:c$.    %  /determined
In~\cite{monferrer1996enhancing}, the same LR calculus is augmented with an interval-based qualitative distance relation of arbitrary granularity.  %  combined/
%~\cite{ESCRIG TOLEDO} combines the same $\mathcal{LR}$ calculus with an interval-based qualitative distance relation of arbitrary granularity.
They encode qualitative spatial constraints in Prolog with CLP (Constraint Logic Programming).
%To check path consistency, the authors implement CHR (Constraint Handling Rules) for composition (constraint propagation) and intersection.
This model is also extended into 3D space~\cite{pacheco2002qualitative}.
%\cite{pacheco2002qualitative} extends this into 3D space.  %  /examines qualitative spatial relations in
%They augment/$\mathcal{LR}$ calculus and Double-Cross calculus are augmented with interval-based distance relation to represent and reason about qualitative position in $\mathbb{R}^3$.

In TPCC calculus \cite{moratz2002qualitative}, LR relations are made finer by further subdividing the 2-D space into four cones. To measure the distance, they draw a circle whose radius is the line segment across reference objects $ab$.
Inside of the circle is designated as \textit{near} and outside of it as \textit{far}.
%This manner/fashion/way, the Eucledean space is partitioned into 24 sections/zones/segments/slices each of which represents/indicate/stands for a distinct/separate orientation- and distance pair/combination.
In another model~\cite{moratz2003spatial,moratzspatial}, the planar space is partitioned into angular segments that are called distance orientation interval (DOI).
%  which they name/         setup
In these calculi, a DOI is specified by four metric parameters $(\phi_1, \phi_2, r_1, r_2)$ and correspond to a qualitative position.

Another approach \cite{moratz2012spatial} for describing qualitative position suggests adding symbolic distance relations into $\mathrm{OPRA}_m$ calculus. The distance relation can be asymmetric 
%i.e. there might be situations in which $a$ is near $b$ but $b$ is far from $a$. Consequently 
hence the distance relation is specified by a pair of relations.
The distance concept is similar to interval-based system \cite{clementini1997qualitative} except that the borders of the intervals also constitute a separate distance relation.

%All the above formalisms consider point objects for describing combination of spatial relations.

Answer Set Programming, thanks to its efficient solvers for computationally hard problems, have been applied to
qualitative spatial reasoning~\cite{baryannis2018trajectory,brenton2016answer,li2012qualitative}. 
% These approaches are monotonic and rely on model checking.   %  constraint satisfaction or
These approaches are based on path consistency and don't involve nonmonotonicity.
As shown in~\cite{Liuetal2010,SkiadKoub2005}, local algorithms such as \textit{path consistency} or \textit{k-consistency} are not sufficient to decide consistency of a CDC network.
%don't generate regions from the domain to check consistency.  %spatial constraints
Encoding of a constraint network in IA and RCC8 has been developed~\cite{brenton2016answer}. % with Answer Set Programming. 
%which implements composition based consistency checking.   %is implemented.
Their formulation can represent disjunctive constraints but not defaults. 
Likewise ASP has been utilized to check path consistency of a network in Trajectory Calculus~\cite{baryannis2018trajectory}.
Unknown relations are nondeterminitically generated and path consistency is tested with a composition table.
In another study, ASP programs for checking consistency of basic and disjunctive constraint networks in any qualitative calculus are presented~\cite{li2012qualitative}. Specialized programs for IA and RCC8 are also provided.   %given
% offers three alternative ASP programs for consistency checking in IA or RCC8.
%The above papers run experiments to judge efficiency of their formulations and compare to other solvers.

%Answer Set Programming, thanks to its efficient solvers for computationally hard problems, has been applied to qualitative spatial reasoning~\cite{brenton2016answer,li2012qualitative}. Encoding of a constraint network in IA or RCC8 is presented by~\cite{brenton2016answer}, % with Answer Set Programming. 
%which implements composition based consistency checking.   %is implemented.
%Their formulation can represent disjunctive constraints but not defaults.  Likewise \cite{baryannis2018trajectory} uses ASP to check path consistency of a network in Trajectory Calculus. Unknown relations are generated and tested with a composition table. ~\cite{li2012qualitative} offers three alternative ASP programs for consistency checking in IA or RCC8. They also run experiments and compare against other solvers to judge efficiency of the programs.

Consistency problems that involve topology (part, whole, contact relations) and 
orientation (left, right, perpendicular, colinear relations) have been solved using ASP Modulo Theories (ASPMT) in~\cite{WalegaBS15}. 
The benefit of ASPMT is that it permits formulas in first order logic and equations including real numbers.
The authors consider point, line segment, circle and polygon as spatial entities.
Constraint networks in Interval Algebra, Rectangle Algebra, LR, RCC8 can be encoded  % $\mathcal{LR}$
in their setting. %, nevertheless Cardinal Directional Calculus of~\cite{Goyal1997, SkiadKoub2004} cannot be encoded. 
Spatial constraints are written in terms of polynomial inequalities in ASPMT and then transformed into SAT Modulo Theories 
for the SMT solver.
%Importantly, in 
For consistency checking in CDC, objects can be instantiated at any shape and size; consequently this approach is not complete for solving the CDC consistency checking problem in this thesis.
Moreover their formulation does not allow for disjunctive, nonmonotonic constraints or preferences.

\section{Our Approach}

\begin{figure}[h]
	\centering
	\begin{tabular}{cc}
		\includegraphics[width=0.3\columnwidth]{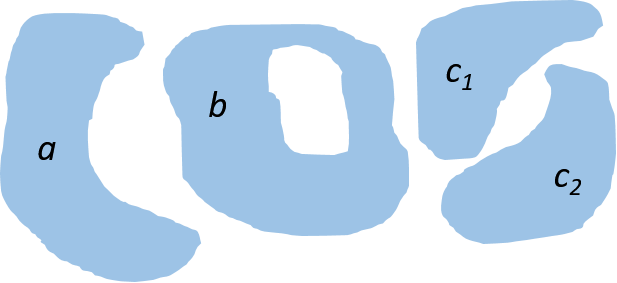} &
		\includegraphics[width=0.36\columnwidth]{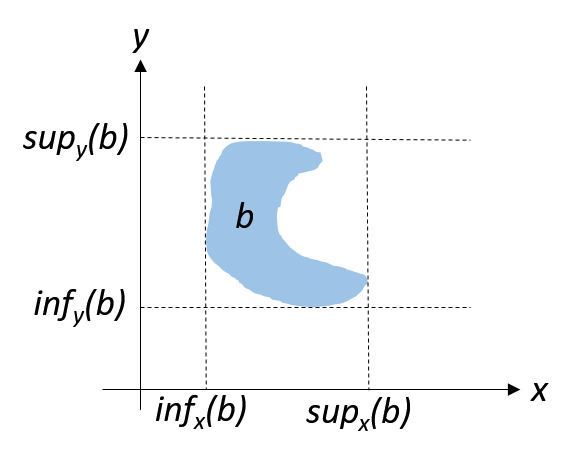}
		\\
		(i) & (ii)
	\end{tabular}
       \\
       \vspace{3mm}
	\begin{tabular}{cc}
		\includegraphics[width=0.27\columnwidth]{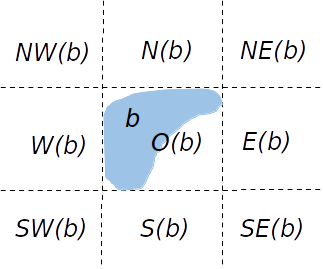} &
		\includegraphics[width=0.45\columnwidth]{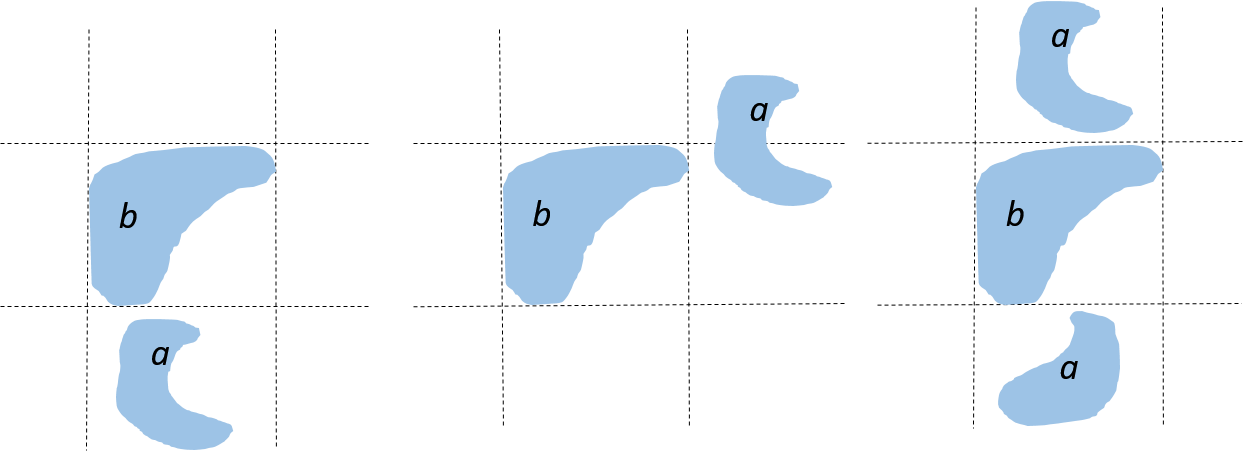}
		\\
		(iii) & (iv)
	\end{tabular}
%	\includegraphics[width=0.32\columnwidth]{regions3.png}
%	\\ (iii) \\
%	\includegraphics[width=0.6\columnwidth]{regions4.png}
%	\\ (iv)
	\caption{(i) Regions $a$, $b$, $c_1$, $c_2$ are connected, where $c = c_1\cup c_2$ is disconnected. (ii) A region and its bounding box.  (iii) Reference tiles. (iv) Sample relations (orientation of $a$ with respect to $b$): $a\ S\ b$, $a\ NE{:}E\ b$, $a\ N{:}S\ b$.}
	\label{fig:regions}
\end{figure}

%To define relative direction of objects with respect to each another, we use Cardinal Directional Calculus (CDC) introduced by~\cite{Goyal1997, SkiadKoub2004}.
We use Cardinal Directional Calculus introduced by Skiadopoulos and Koubarakis~\cite{Goyal1997, SkiadKoub2004} to define relative direction of objects with respect to each other.
In CDC, direction between objects are denoted by binary relations.
Objects are extended regions on a plane and they can be \textit{simple}, \textit{connected} or \textit{disconnected} as shown in Figure~\ref{fig:regions}(i).
The minimum bounding rectangle of the reference object along the axes (Figure~\ref{fig:regions}(ii)) divides the plane into nine regions (called tiles): 
%$O(b)$, $S(b)$, $SW(b)$, $W(b)$, $NW(b)$, $N(b)$, $NE(b)$, $E(b)$, $SE(b)$ (Fig.~\ref{fig:regions}(iii)). 
\textit{north} (N), \textit{south} (S), \textit{east} (E), \textit{west} (W), \textit{northeast} (NE), \textit{northwest} (NW), \textit{southeast} (SE), \textit{southwest} (SW), \textit{on} (O) as in Figure~\ref{fig:regions}(iii).    % (Figure~\ref{fig:regions}(iii)).
These nine atomic (single-tile) relations and their combinations constitute the set of basic relations. (e.g. see Figure~\ref{fig:regions}(iv))
CDC also allows for disjunction of these basic relations.  %disjunctive relations.    % disjunction of basic relations.

\vspace{3mm}

Using these binary relations, relative directions of extended objects can be described in CDC as a set of constraints.
In our studies, we formalize CDC using ASP and further extend it with a new form of constraints: Default CDC constraints.
They can be used to express default assumptions like ``The food truck is normally to the south of the movie theater".

We define qualitative distance relations with adjustable granularity $g_d$.
To examplify, for granularity $g_d=6$, the set of basic distance relations are 
%$ \Omega \eqs $ \textit{\{adjacent, very near, near, commensurate, far, very far \}.}
$ \Omega \eqs \mathit{ \{adjacent, \: very \: near, \: near, \: commensurate, \: far},$ \\ $ \mathit{very \: far \} }$.

%In our proposed calculus, we combine direction and distance to define qualitative position of non-point objects on the plane.
%The qualitative distance relation we use is novel. 
%It is a binary relation that can take finite number of symbolic values such as   % The distance relation between objects
%\textit{adjacent}, \textit{near}, \textit{commensurate}, \textit{far} whose granularity can be increased when necessary.

One of the central problems in CDC and qualitative spatial reasoning literature is the consistency checking of a constraint network.
The input of the consistency checking problem are a set of spatial variables (objects), the domain of objects, a constraint network and
the set of CDC relations. The domain can be % \textbf{\textit{Reg}}   \textbf{\textit{Reg*}} (
the set of connected regions or 
the set of possibly disconnected regions in $\mathbb{R}^2$.
Then, the consistency checking problem asks for whether there exists an instantiation of
objects in the domain which satisfy all constraints in the network. If such an instantiation exists,
the output is \ii{Yes}, otherwise it is \ii{No}.

Based on our representation of direction and distance constraints in ASP, we propose a novel method to check consistency of a network of constraints. Note that consistency checking problem is defined over continuous domain.
We discretize consistency checking problem in CDC, prove its equivalence with the continuous version, and introduce a solution using ASP.
We also establish soundness and completeness of our ASP-based solution.
Namely, the ASP program has an answer set if and only if the given network of constraints is consistent.

Our ASP formulation is elaboration tolerant in the sense that a few rules are added to the main ASP program in order to   % concatenated
incorporate disjunctive, default, soft, negative constraints, or to ensure that the generated regions are connected.   % or test whether

\section{Goal and Current Status of the Research}    %  OBJECTIVES OF THE THESIS

%Our objective in this thesis is to endow the agent with qualitative representation of orientation
%so that it can inspect consistency of given information, infer unknown spatial relations, incorporate commonsense knowledge into reasoning
%and deal with indefinite or incomplete information.

The objective of this thesis is to develop a generic formal framework to represent and reason about qualitative spatial relations. 
In the first step, we have studied directional relations and taken Cardinal Directional Calculus as a starting point.
We have formulated CDC consistency checking problem in ASP. Then we have extended CDC with new sorts of constraints which involve defaults, preferences, negation using ASP \cite{izmirlioglu018}. We call this extended version of CDC as nonmonotonic CDC (nCDC). 

Currently, we are working on a further extension of nCDC with qualitative distance relation. We name this extension as nCDC+. Preferences, disjunctive and default constraints can also be expressed in nCDC+.

%Next, in order to represent and reason with multiple aspects we further extend nCDC with qualitative distance relation for an integrated qualitative calculus and name the new calculus as nCDC+. Preferences, disjunctive and default constraints can also be expressed in nCDC+.

For CDC, nCDC, nCDC+, we aim to introduce a general framework to solve consistency checking problems,
address composition and inversion of qualitative spatial relations, infer unknown or missing relations between objects, and
find a suitable configuration of objects which fulfills the given spatial constraints in the inquiry.

We have illustrated benefits of our methods for reasoning over nCDC constraints in \cite{izmirlioglu018} with the example scenarios mentioned in the introduction. 
%These scenarios show how ASP formulation can be used for deciding consistency, handling indefinite information, nonmonothonic reasoning and inferring unknown relations.
We have evaluated efficiency of our approach for consistency checking in nCDC with experiments on benchmark instances.
For this purpose, a variety of problem instances over differents domain have been prepared. % for testing consistency
%For each domain, there are groups of instances containing disjunctive constraints, default constraints and incomplete networks.
For every instance, grounding time, total computation time and program size have been recorded. Observed values are compared across input parameters and the domain. We plan to perform these applications and experiments for nCDC+ as well.

\vspace{0.1in}

\section{Future Work}

Our agenda for future work consists of the following items:\\
\begin{itemize}
%\item \textbf{Experimental evaluation for nCDC :} Create benchmark instances with negative and soft CDC constraints, run experiments to assess the impact of input size and existence of negative/soft information on performance. 
\item \textbf{Experimental evaluation for nCDC+ :} We plan to create benchmark instances with nCDC+ networks that include directional, distance constraints, run experiments and evaluate the results with respect to computation time. 
\item \textbf{Applications of nCDC+:} We plan to revise example scenarios in the introduction for nCDC+ and apply our ASP-based methods to solve them.
%\item \textbf{Geometric rearrangement problem:} Discretization of the problem and theoretical analysis of problem size, developing ASP programs that embed external engines for low-level collision check, creating benchmark instances and empirical evaluation of our approach.  
% with experiments.
\end{itemize}

%\nocite{*}
\bibliographystyle{eptcs}
\bibliography{referencesAugust2018}

\begin{thebibliography}{10}
\providecommand{\bibitemdeclare}[2]{}
\providecommand{\surnamestart}{}
\providecommand{\surnameend}{}
\providecommand{\urlprefix}{Available at }
\providecommand{\url}[1]{\texttt{#1}}
\providecommand{\href}[2]{\texttt{#2}}
\providecommand{\urlalt}[2]{\href{#1}{#2}}
\providecommand{\doi}[1]{doi:\urlalt{http://dx.doi.org/#1}{#1}}
\providecommand{\bibinfo}[2]{#2}

\bibitemdeclare{article}{Allen83}
\bibitem{Allen83}
\bibinfo{author}{James~F. \surnamestart Allen\surnameend}
  (\bibinfo{year}{1983}): \emph{\bibinfo{title}{Maintaining Knowledge about
  Temporal Intervals}}.
\newblock {\sl \bibinfo{journal}{Commun. {ACM}}}
  \bibinfo{volume}{26}(\bibinfo{number}{11}), pp. \bibinfo{pages}{832--843},
  \doi{10.1145/182.358434}.

\bibitemdeclare{inproceedings}{balbiani1998model}
\bibitem{balbiani1998model}
\bibinfo{author}{Philippe \surnamestart Balbiani\surnameend},
  \bibinfo{author}{Jean-Fran{\c{c}}ois \surnamestart Condotta\surnameend} \&
  \bibinfo{author}{Luis~Fari{\~n}as \surnamestart del Cerro\surnameend}
  (\bibinfo{year}{1998}): \emph{\bibinfo{title}{A model for reasoning about
  bidemsional temporal relations}}.
\newblock In: {\sl \bibinfo{booktitle}{Proceedings of the Sixth International
  Conference on Principles of Knowledge Representation and Reasoning}},
  \bibinfo{organization}{Morgan Kaufmann Publishers Inc.}, pp.
  \bibinfo{pages}{124--130}.

\bibitemdeclare{inproceedings}{BalbianiCC99}
\bibitem{BalbianiCC99}
\bibinfo{author}{Philippe \surnamestart Balbiani\surnameend},
  \bibinfo{author}{Jean{-}Fran{\c{c}}ois \surnamestart Condotta\surnameend} \&
  \bibinfo{author}{Luis~Fari{\~{n}}as \surnamestart del Cerro\surnameend}
  (\bibinfo{year}{1999}): \emph{\bibinfo{title}{A New Tractable Subclass of the
  Rectangle Algebra}}.
\newblock In: {\sl \bibinfo{booktitle}{Proc. of IJCAI}}, pp.
  \bibinfo{pages}{442--447}.

\bibitemdeclare{article}{baryannis2018trajectory}
\bibitem{baryannis2018trajectory}
\bibinfo{author}{George \surnamestart Baryannis\surnameend},
  \bibinfo{author}{Ilias \surnamestart Tachmazidis\surnameend},
  \bibinfo{author}{Sotiris \surnamestart Batsakis\surnameend},
  \bibinfo{author}{Grigoris \surnamestart Antoniou\surnameend},
  \bibinfo{author}{Mario \surnamestart Alviano\surnameend},
  \bibinfo{author}{Timos \surnamestart Sellis\surnameend} \&
  \bibinfo{author}{Pei-Wei \surnamestart Tsai\surnameend}
  (\bibinfo{year}{2018}): \emph{\bibinfo{title}{A Trajectory Calculus for
  Qualitative Spatial Reasoning Using Answer Set Programming}}.
\newblock {\sl \bibinfo{journal}{Theory and Practice of Logic Programming}}
  \bibinfo{volume}{18}(\bibinfo{number}{3-4}), pp. \bibinfo{pages}{355--371},
  \doi{10.1007/s10707-007-0023-2}.

\bibitemdeclare{inproceedings}{brageul2007model}
\bibitem{brageul2007model}
\bibinfo{author}{David \surnamestart Brageul\surnameend} \&
  \bibinfo{author}{Hans~W \surnamestart Guesgen\surnameend}
  (\bibinfo{year}{2007}): \emph{\bibinfo{title}{A Model for Qualitative Spatial
  Reasoning Combining Topology, Orientation and Distance.}}
\newblock In: {\sl \bibinfo{booktitle}{FLAIRS Conference}}, pp.
  \bibinfo{pages}{653--658}.

\bibitemdeclare{inproceedings}{brenton2016answer}
\bibitem{brenton2016answer}
\bibinfo{author}{Christopher \surnamestart Brenton\surnameend},
  \bibinfo{author}{Wolfgang \surnamestart Faber\surnameend} \&
  \bibinfo{author}{Sotiris \surnamestart Batsakis\surnameend}
  (\bibinfo{year}{2016}): \emph{\bibinfo{title}{Answer Set Programming for
  Qualitative Spatio-Temporal Reasoning: Methods and Experiments}}.
\newblock In: {\sl \bibinfo{booktitle}{OASIcs-OpenAccess Series in
  Informatics}}, \bibinfo{volume}{52}, \bibinfo{organization}{Schloss
  Dagstuhl-Leibniz-Zentrum fuer Informatik}.

\bibitemdeclare{article}{chen2015survey}
\bibitem{chen2015survey}
\bibinfo{author}{Juan \surnamestart Chen\surnameend},
  \bibinfo{author}{Anthony~G \surnamestart Cohn\surnameend},
  \bibinfo{author}{Dayou \surnamestart Liu\surnameend},
  \bibinfo{author}{Shengsheng \surnamestart Wang\surnameend},
  \bibinfo{author}{Jihong \surnamestart Ouyang\surnameend} \&
  \bibinfo{author}{Qiangyuan \surnamestart Yu\surnameend}
  (\bibinfo{year}{2015}): \emph{\bibinfo{title}{A survey of qualitative spatial
  representations}}.
\newblock {\sl \bibinfo{journal}{The Knowledge Engineering Review}}
  \bibinfo{volume}{30}(\bibinfo{number}{1}), pp. \bibinfo{pages}{106--136},
  \doi{10.1016/j.artint.2011.10.003}.

\bibitemdeclare{article}{clementini1997qualitative}
\bibitem{clementini1997qualitative}
\bibinfo{author}{Eliseo \surnamestart Clementini\surnameend},
  \bibinfo{author}{Paolino \surnamestart Di~Felice\surnameend} \&
  \bibinfo{author}{Daniel \surnamestart Hern{\'a}ndez\surnameend}
  (\bibinfo{year}{1997}): \emph{\bibinfo{title}{Qualitative representation of
  positional information}}.
\newblock {\sl \bibinfo{journal}{Artificial intelligence}}
  \bibinfo{volume}{95}(\bibinfo{number}{2}), pp. \bibinfo{pages}{317--356},
  \doi{10.1016/S0004-3702(97)00046-5}.

\bibitemdeclare{article}{cohn1997qualitative}
\bibitem{cohn1997qualitative}
\bibinfo{author}{Anthony~G \surnamestart Cohn\surnameend},
  \bibinfo{author}{Brandon \surnamestart Bennett\surnameend},
  \bibinfo{author}{John \surnamestart Gooday\surnameend} \&
  \bibinfo{author}{Nicholas~Mark \surnamestart Gotts\surnameend}
  (\bibinfo{year}{1997}): \emph{\bibinfo{title}{Qualitative spatial
  representation and reasoning with the region connection calculus}}.
\newblock {\sl \bibinfo{journal}{GeoInformatica}}
  \bibinfo{volume}{1}(\bibinfo{number}{3}), pp. \bibinfo{pages}{275--316},
  \doi{10.1023/A:1009712514511}.

\bibitemdeclare{article}{cohn2008qualitative}
\bibitem{cohn2008qualitative}
\bibinfo{author}{Anthony~G \surnamestart Cohn\surnameend} \&
  \bibinfo{author}{Jochen \surnamestart Renz\surnameend}
  (\bibinfo{year}{2008}): \emph{\bibinfo{title}{Qualitative Spatial
  Representation and Reasoning}}.
\newblock {\sl \bibinfo{journal}{Handbook of Knowledge Representation}}, p.
  \bibinfo{pages}{551}, \doi{10.1016/S1574-6526(07)03013-1}.

\bibitemdeclare{article}{dorr2014qualitative}
\bibitem{dorr2014qualitative}
\bibinfo{author}{Christopher~H \surnamestart Dorr\surnameend} \&
  \bibinfo{author}{Reinhard \surnamestart Moratz\surnameend}
  (\bibinfo{year}{2014}): \emph{\bibinfo{title}{Qualitative shape
  representation based on the qualitative relative direction and distance
  calculus eOPRAm}}.
\newblock {\sl \bibinfo{journal}{arXiv preprint arXiv:1412.6649}}.

\bibitemdeclare{inproceedings}{dugat1999qualitative}
\bibitem{dugat1999qualitative}
\bibinfo{author}{Vincent \surnamestart Dugat\surnameend},
  \bibinfo{author}{Pierre \surnamestart Gambarotto\surnameend} \&
  \bibinfo{author}{Yannick \surnamestart Larvor\surnameend}
  (\bibinfo{year}{1999}): \emph{\bibinfo{title}{Qualitative theory of shape and
  orientation}}.
\newblock In: {\sl \bibinfo{booktitle}{Proc. of the 16th Int. Joint Conference
  on Artificial Intelligence (IJCAI'99), Stockolm, Sweden}}, pp.
  \bibinfo{pages}{45--53}.

\bibitemdeclare{article}{dylla2017survey}
\bibitem{dylla2017survey}
\bibinfo{author}{Frank \surnamestart Dylla\surnameend},
  \bibinfo{author}{Jae~Hee \surnamestart Lee\surnameend}, \bibinfo{author}{Till
  \surnamestart Mossakowski\surnameend}, \bibinfo{author}{Thomas \surnamestart
  Schneider\surnameend}, \bibinfo{author}{Andr{\'e}~Van \surnamestart
  Delden\surnameend}, \bibinfo{author}{Jasper Van~De \surnamestart
  Ven\surnameend} \& \bibinfo{author}{Diedrich \surnamestart Wolter\surnameend}
  (\bibinfo{year}{2017}): \emph{\bibinfo{title}{A survey of qualitative spatial
  and temporal calculi: algebraic and computational properties}}.
\newblock {\sl \bibinfo{journal}{ACM Computing Surveys (CSUR)}}
  \bibinfo{volume}{50}(\bibinfo{number}{1}), p.~\bibinfo{pages}{7},
  \doi{10.1145/3038927}.

\bibitemdeclare{article}{egenhofer1990categorizing}
\bibitem{egenhofer1990categorizing}
\bibinfo{author}{Max~J \surnamestart Egenhofer\surnameend} \&
  \bibinfo{author}{John \surnamestart Herring\surnameend}
  (\bibinfo{year}{1990}): \emph{\bibinfo{title}{Categorizing binary topological
  relations between regions, lines, and points in geographic databases}}.
\newblock {\sl \bibinfo{journal}{The}}
  \bibinfo{volume}{9}(\bibinfo{number}{94-1}), p.~\bibinfo{pages}{76}.

\bibitemdeclare{article}{falomir2013qualitative}
\bibitem{falomir2013qualitative}
\bibinfo{author}{Zoe \surnamestart Falomir\surnameend},
  \bibinfo{author}{Lled{\'o} \surnamestart Museros\surnameend},
  \bibinfo{author}{Vicent \surnamestart Castell{\'o}\surnameend} \&
  \bibinfo{author}{Luis \surnamestart Gonzalez-Abril\surnameend}
  (\bibinfo{year}{2013}): \emph{\bibinfo{title}{Qualitative distances and
  qualitative image descriptions for representing indoor scenes in robotics}}.
\newblock {\sl \bibinfo{journal}{Pattern Recognition Letters}}
  \bibinfo{volume}{34}(\bibinfo{number}{7}), pp. \bibinfo{pages}{731--743},
  \doi{10.1016/j.patrec.2012.08.012}.

\bibitemdeclare{inproceedings}{Frank91}
\bibitem{Frank91}
\bibinfo{author}{A.~U. \surnamestart Frank\surnameend} (\bibinfo{year}{1991}):
  \emph{\bibinfo{title}{Qualitative Spatial Reasoning about Cardinal
  Directions}}.
\newblock In: {\sl \bibinfo{booktitle}{Proc. of Auto-Carto 10}}.

\bibitemdeclare{inbook}{Freksa1992}
\bibitem{Freksa1992}
\bibinfo{author}{Christian \surnamestart Freksa\surnameend}
  (\bibinfo{year}{1992}): \emph{\bibinfo{title}{Using orientation information
  for qualitative spatial reasoning}}, pp. \bibinfo{pages}{162--178}.
\newblock \bibinfo{publisher}{Springer Berlin Heidelberg}.

\bibitemdeclare{article}{gerevini2002combining}
\bibitem{gerevini2002combining}
\bibinfo{author}{Alfonso \surnamestart Gerevini\surnameend} \&
  \bibinfo{author}{Jochen \surnamestart Renz\surnameend}
  (\bibinfo{year}{2002}): \emph{\bibinfo{title}{Combining topological and size
  information for spatial reasoning}}.
\newblock {\sl \bibinfo{journal}{Artificial Intelligence}}
  \bibinfo{volume}{137}(\bibinfo{number}{1-2}), pp. \bibinfo{pages}{1--42},
  \doi{10.1016/S0004-3702(02)00193-5}.

\bibitemdeclare{article}{gottfried2005global}
\bibitem{gottfried2005global}
\bibinfo{author}{Bj{\"o}rn \surnamestart Gottfried\surnameend}
  (\bibinfo{year}{2005}): \emph{\bibinfo{title}{Global feature schemes for
  qualitative shape descriptions}}.
\newblock {\sl \bibinfo{journal}{IJCAI-05 WS on spatial and temporal
  reasoning}}.

\bibitemdeclare{article}{Goyal1997}
\bibitem{Goyal1997}
\bibinfo{author}{R~\surnamestart Goyal\surnameend} \& \bibinfo{author}{Max~J
  \surnamestart Egenhofer\surnameend} (\bibinfo{year}{1997}):
  \emph{\bibinfo{title}{The direction-relation matrix: A representation for
  directions relations between extended spatial objects}}.
\newblock {\sl \bibinfo{journal}{The annual assembly and the summer retreat of
  University Consortium for Geographic Information Systems Science}}
  \bibinfo{volume}{3}, pp. \bibinfo{pages}{95--102}.

\bibitemdeclare{article}{guesgen2002reasoning}
\bibitem{guesgen2002reasoning}
\bibinfo{author}{Hans~W \surnamestart Guesgen\surnameend}
  (\bibinfo{year}{2002}): \emph{\bibinfo{title}{Reasoning about distance based
  on fuzzy sets}}.
\newblock {\sl \bibinfo{journal}{Applied Intelligence}}
  \bibinfo{volume}{17}(\bibinfo{number}{3}), pp. \bibinfo{pages}{265--270},
  \doi{10.1023/A:1020024013757}.

\bibitemdeclare{inproceedings}{hong1995robustness}
\bibitem{hong1995robustness}
\bibinfo{author}{Jung-Hong \surnamestart Hong\surnameend}, \bibinfo{author}{Max
  \surnamestart Egenhofer\surnameend} \& \bibinfo{author}{Andrew~U
  \surnamestart Frank\surnameend} (\bibinfo{year}{1995}):
  \emph{\bibinfo{title}{On the robustness of qualitative distance-and
  direction-reasoning}}.
\newblock In: {\sl \bibinfo{booktitle}{Autocarto Conference}}, pp.
  \bibinfo{pages}{301--310}.

\bibitemdeclare{article}{hotzcombining}
\bibitem{hotzcombining}
\bibinfo{author}{Lothar \surnamestart Hotz\surnameend}, \bibinfo{author}{Pascal
  \surnamestart Rost\surnameend} \& \bibinfo{author}{Stephanie \surnamestart
  von Riegen\surnameend}: \emph{\bibinfo{title}{Combining Qualitative Spatial
  Reasoning and Ontological Reasoning for Supporting Robot Tasks}}.

\bibitemdeclare{inproceedings}{izmirlioglu018}
\bibitem{izmirlioglu018}
\bibinfo{author}{Yusuf \surnamestart Izmirlioglu\surnameend} \&
  \bibinfo{author}{Esra \surnamestart Erdem\surnameend} (\bibinfo{year}{2018}):
  \emph{\bibinfo{title}{Qualitative Reasoning About Cardinal Directions Using
  Answer Set Programming}}.
\newblock In: {\sl \bibinfo{booktitle}{Proc. of AAAI}}.

\bibitemdeclare{inproceedings}{LeeRW13}
\bibitem{LeeRW13}
\bibinfo{author}{Jae~Hee \surnamestart Lee\surnameend}, \bibinfo{author}{Jochen
  \surnamestart Renz\surnameend} \& \bibinfo{author}{Diedrich \surnamestart
  Wolter\surnameend} (\bibinfo{year}{2013}): \emph{\bibinfo{title}{StarVars -
  Effective Reasoning about Relative Directions}}.
\newblock In: {\sl \bibinfo{booktitle}{Proc. of IJCAI}}, pp.
  \bibinfo{pages}{976--982}.

\bibitemdeclare{inproceedings}{li2012qualitative}
\bibitem{li2012qualitative}
\bibinfo{author}{Jason~Jingshi \surnamestart Li\surnameend}
  (\bibinfo{year}{2012}): \emph{\bibinfo{title}{Qualitative spatial and
  temporal reasoning with answer set programming}}.
\newblock In: {\sl \bibinfo{booktitle}{Tools with Artificial Intelligence
  (ICTAI), 2012 IEEE 24th International Conference on}}, \bibinfo{volume}{1},
  \bibinfo{organization}{IEEE}, pp. \bibinfo{pages}{603--609},
  \doi{10.1109/ICTAI.2012.87}.

\bibitemdeclare{inproceedings}{ligozat1993qualitative}
\bibitem{ligozat1993qualitative}
\bibinfo{author}{G{\'e}rard~F \surnamestart Ligozat\surnameend}
  (\bibinfo{year}{1993}): \emph{\bibinfo{title}{Qualitative triangulation for
  spatial reasoning}}.
\newblock In: {\sl \bibinfo{booktitle}{European Conference on Spatial
  Information Theory}}, \bibinfo{organization}{Springer}, pp.
  \bibinfo{pages}{54--68}.

\bibitemdeclare{phdthesis}{Liuthesis2013}
\bibitem{Liuthesis2013}
\bibinfo{author}{Weiming \surnamestart Liu\surnameend} (\bibinfo{year}{2013}):
  \emph{\bibinfo{title}{Qualitative constraint satisfaction problems:
  algorithms, computational complexity, and extended framework}}.
\newblock Ph.D. thesis, \bibinfo{school}{University of Technology, Sydney}.

\bibitemdeclare{article}{LiLiu2011}
\bibitem{LiLiu2011}
\bibinfo{author}{Weiming \surnamestart Liu\surnameend} \&
  \bibinfo{author}{Sanjiang \surnamestart Li\surnameend}
  (\bibinfo{year}{2011}): \emph{\bibinfo{title}{Reasoning about cardinal
  directions between extended objects: The NP-hardness result}}.
\newblock {\sl \bibinfo{journal}{Artificial Intelligence}}
  \bibinfo{volume}{175}(\bibinfo{number}{18}), pp. \bibinfo{pages}{2155--2169},
  \doi{10.1016/j.artint.2011.07.005}.

\bibitemdeclare{article}{Liuetal2010}
\bibitem{Liuetal2010}
\bibinfo{author}{Weiming \surnamestart Liu\surnameend},
  \bibinfo{author}{Xiaotong \surnamestart Zhang\surnameend},
  \bibinfo{author}{Sanjiang \surnamestart Li\surnameend} \&
  \bibinfo{author}{Mingsheng \surnamestart Ying\surnameend}
  (\bibinfo{year}{2010}): \emph{\bibinfo{title}{Reasoning about cardinal
  directions between extended objects}}.
\newblock {\sl \bibinfo{journal}{Artificial Intelligence}}
  \bibinfo{volume}{174}(\bibinfo{number}{12-13}), pp.
  \bibinfo{pages}{951--983}, \doi{10.1016/j.artint.2010.05.006}.

\bibitemdeclare{inproceedings}{monferrer1996enhancing}
\bibitem{monferrer1996enhancing}
\bibinfo{author}{M~Teresa~Escrig \surnamestart Monferrer\surnameend} \&
  \bibinfo{author}{Francisco~Toledo \surnamestart Lobo\surnameend}
  (\bibinfo{year}{1996}): \emph{\bibinfo{title}{Enhancing qualitative relative
  orientation with qualitative distance for robot path planning}}.
\newblock In: {\sl \bibinfo{booktitle}{Tools with Artificial Intelligence,
  1996., Proceedings Eighth IEEE International Conference on}},
  \bibinfo{organization}{IEEE}, pp. \bibinfo{pages}{174--182}.

\bibitemdeclare{inproceedings}{moratz2005relative}
\bibitem{moratz2005relative}
\bibinfo{author}{Reinhard \surnamestart Moratz\surnameend},
  \bibinfo{author}{Frank \surnamestart Dylla\surnameend} \&
  \bibinfo{author}{Lutz \surnamestart Frommberger\surnameend}
  (\bibinfo{year}{2005}): \emph{\bibinfo{title}{A relative orientation algebra
  with adjustable granularity}}.
\newblock In: {\sl \bibinfo{booktitle}{Proceedings of the Workshop on Agents in
  Real-Time and Dynamic Environments (IJCAI 05)}}, \bibinfo{volume}{21},
  p.~\bibinfo{pages}{22}.

\bibitemdeclare{inproceedings}{moratz2002qualitative}
\bibitem{moratz2002qualitative}
\bibinfo{author}{Reinhard \surnamestart Moratz\surnameend},
  \bibinfo{author}{Bernhard \surnamestart Nebel\surnameend} \&
  \bibinfo{author}{Christian \surnamestart Freksa\surnameend}
  (\bibinfo{year}{2002}): \emph{\bibinfo{title}{Qualitative spatial reasoning
  about relative position}}.
\newblock In: {\sl \bibinfo{booktitle}{International Conference on Spatial
  Cognition}}, \bibinfo{organization}{Springer}, pp. \bibinfo{pages}{385--400}.

\bibitemdeclare{inproceedings}{moratz2000qualitative}
\bibitem{moratz2000qualitative}
\bibinfo{author}{Reinhard \surnamestart Moratz\surnameend},
  \bibinfo{author}{Jochen \surnamestart Renz\surnameend} \&
  \bibinfo{author}{Diedrich \surnamestart Wolter\surnameend}
  (\bibinfo{year}{2000}): \emph{\bibinfo{title}{Qualitative spatial reasoning
  about line segments}}.
\newblock In: {\sl \bibinfo{booktitle}{ECAI}}, pp. \bibinfo{pages}{234--238}.

\bibitemdeclare{incollection}{moratzspatial}
\bibitem{moratzspatial}
\bibinfo{author}{Reinhard \surnamestart Moratz\surnameend} \&
  \bibinfo{author}{Jan~Oliver \surnamestart Wallgr{\"u}n\surnameend}:
  \emph{\bibinfo{title}{Spatial Reasoning about Relative Orientation and
  Distance for Robot Exploration}}.
\newblock In: {\sl \bibinfo{booktitle}{Spatial Information Theory. Foundations
  of Geographic Information Science}}, \doi{10.1007/BF00117601}.

\bibitemdeclare{inproceedings}{moratz2003spatial}
\bibitem{moratz2003spatial}
\bibinfo{author}{Reinhard \surnamestart Moratz\surnameend} \&
  \bibinfo{author}{Jan~Oliver \surnamestart Wallgr{\"u}n\surnameend}
  (\bibinfo{year}{2003}): \emph{\bibinfo{title}{Spatial reasoning about
  relative orientation and distance for robot exploration}}.
\newblock In: {\sl \bibinfo{booktitle}{International Conference on Spatial
  Information Theory}}, \bibinfo{organization}{Springer}, pp.
  \bibinfo{pages}{61--74}.

\bibitemdeclare{article}{moratz2012spatial}
\bibitem{moratz2012spatial}
\bibinfo{author}{Reinhard \surnamestart Moratz\surnameend} \&
  \bibinfo{author}{Jan~Oliver \surnamestart Wallgr{\"u}n\surnameend}
  (\bibinfo{year}{2012}): \emph{\bibinfo{title}{Spatial reasoning with
  augmented points: Extending cardinal directions with local distances}}.
\newblock {\sl \bibinfo{journal}{Journal of Spatial Information Science}}
  \bibinfo{volume}{2012}(\bibinfo{number}{5}), pp. \bibinfo{pages}{1--30}.

\bibitemdeclare{inproceedings}{museros2004qualitative}
\bibitem{museros2004qualitative}
\bibinfo{author}{Lled{\'o} \surnamestart Museros\surnameend} \&
  \bibinfo{author}{M~Teresa \surnamestart Escrig\surnameend}
  (\bibinfo{year}{2004}): \emph{\bibinfo{title}{A qualitative theory for shape
  representation and matching for design.}}
\newblock In: {\sl \bibinfo{booktitle}{Proceedings of the 16th European
  Conference on Artificial Intelligence}}, \bibinfo{organization}{IOS Press},
  pp. \bibinfo{pages}{858--862}.

\bibitemdeclare{inproceedings}{Navarreteetal2007}
\bibitem{Navarreteetal2007}
\bibinfo{author}{Isabel \surnamestart Navarrete\surnameend},
  \bibinfo{author}{Antonio \surnamestart Morales\surnameend} \&
  \bibinfo{author}{Guido \surnamestart Sciavicco\surnameend}
  (\bibinfo{year}{2007}): \emph{\bibinfo{title}{Consistency Checking of Basic
  Cardinal Constraints over Connected Regions}}.
\newblock In: {\sl \bibinfo{booktitle}{Proc. of IJCAI}}, pp.
  \bibinfo{pages}{495--500}.
\newblock
  \urlprefix\url{http://dli.iiit.ac.in/ijcai/IJCAI-2007/PDF/IJCAI07-078.pdf}.

\bibitemdeclare{inproceedings}{pacheco2002qualitative}
\bibitem{pacheco2002qualitative}
\bibinfo{author}{Julio \surnamestart Pacheco\surnameend},
  \bibinfo{author}{M{\textordfeminine}~Teresa \surnamestart Escrig\surnameend}
  \& \bibinfo{author}{Francisco \surnamestart Toledo\surnameend}
  (\bibinfo{year}{2002}): \emph{\bibinfo{title}{Qualitative spatial reasoning
  on three-dimensional orientation point objects}}.
\newblock In: {\sl \bibinfo{booktitle}{Proccedings of the QR2002. 16th
  International WorkShop on Qualitative Reasoning. Editors: Nuria Agell and}}.

\bibitemdeclare{article}{SkiadKoub2004}
\bibitem{SkiadKoub2004}
\bibinfo{author}{Spiros \surnamestart Skiadopoulos\surnameend} \&
  \bibinfo{author}{Manolis \surnamestart Koubarakis\surnameend}
  (\bibinfo{year}{2004}): \emph{\bibinfo{title}{Composing cardinal direction
  relations}}.
\newblock {\sl \bibinfo{journal}{Artificial Intelligence}}
  \bibinfo{volume}{152}(\bibinfo{number}{2}), pp. \bibinfo{pages}{143--171},
  \doi{10.1016/S0004-3702(03)00137-1}.

\bibitemdeclare{article}{SkiadKoub2005}
\bibitem{SkiadKoub2005}
\bibinfo{author}{Spiros \surnamestart Skiadopoulos\surnameend} \&
  \bibinfo{author}{Manolis \surnamestart Koubarakis\surnameend}
  (\bibinfo{year}{2005}): \emph{\bibinfo{title}{On the consistency of cardinal
  direction constraints}}.
\newblock {\sl \bibinfo{journal}{Artificial Intelligence}}
  \bibinfo{volume}{163}(\bibinfo{number}{1}), pp. \bibinfo{pages}{91--135},
  \doi{10.1016/j.artint.2004.10.010}.

\bibitemdeclare{inproceedings}{WalegaBS15}
\bibitem{WalegaBS15}
\bibinfo{author}{Przemyslaw~Andrzej \surnamestart Walega\surnameend},
  \bibinfo{author}{Mehul \surnamestart Bhatt\surnameend} \&
  \bibinfo{author}{Carl P.~L. \surnamestart Schultz\surnameend}
  (\bibinfo{year}{2015}): \emph{\bibinfo{title}{{ASPMT(QS):} Non-Monotonic
  Spatial Reasoning with Answer Set Programming Modulo Theories}}.
\newblock In: {\sl \bibinfo{booktitle}{Proc. of LPNMR}}, pp.
  \bibinfo{pages}{488--501}.

\bibitemdeclare{inproceedings}{van2005double}
\bibitem{van2005double}
\bibinfo{author}{Nico \surnamestart Van~de Weghe\surnameend},
  \bibinfo{author}{Guy \surnamestart De~Tr{\'e}\surnameend},
  \bibinfo{author}{Bart \surnamestart Kuijpers\surnameend} \&
  \bibinfo{author}{Philippe \surnamestart De~Maeyer\surnameend}
  (\bibinfo{year}{2005}): \emph{\bibinfo{title}{The double-cross and the
  generalization concept as a basis for representing and comparing shapes of
  polylines}}.
\newblock In: {\sl \bibinfo{booktitle}{OTM Confederated International
  Conferences" On the Move to Meaningful Internet Systems"}},
  \bibinfo{organization}{Springer}, pp. \bibinfo{pages}{1087--1096}.

\bibitemdeclare{inproceedings}{Zhangetal2008}
\bibitem{Zhangetal2008}
\bibinfo{author}{Xiaotong \surnamestart Zhang\surnameend},
  \bibinfo{author}{Weiming \surnamestart Liu\surnameend},
  \bibinfo{author}{Sanjiang \surnamestart Li\surnameend} \&
  \bibinfo{author}{Mingsheng \surnamestart Ying\surnameend}
  (\bibinfo{year}{2008}): \emph{\bibinfo{title}{Reasoning with Cardinal
  Directions: An Efficient Algorithm}}.
\newblock In: {\sl \bibinfo{booktitle}{Proceedings of the Twenty-Third {AAAI}
  Conference on Artificial Intelligence, {AAAI} 2008, Chicago, Illinois, USA,
  July 13-17, 2008}}, pp. \bibinfo{pages}{387--392}.

\bibitemdeclare{article}{zimmermann1996qualitative}
\bibitem{zimmermann1996qualitative}
\bibinfo{author}{Kai \surnamestart Zimmermann\surnameend} \&
  \bibinfo{author}{Christian \surnamestart Freksa\surnameend}
  (\bibinfo{year}{1996}): \emph{\bibinfo{title}{Qualitative spatial reasoning
  using orientation, distance, and path knowledge}}.
\newblock {\sl \bibinfo{journal}{Applied intelligence}}
  \bibinfo{volume}{6}(\bibinfo{number}{1}), pp. \bibinfo{pages}{49--58},
  \doi{10.1007/BF00117601}.

\end{thebibliography}
\end{document}